\pdfoutput=1

\documentclass[11pt]{article}

\usepackage{ACL2023}

\usepackage{times}
\usepackage{latexsym}

\usepackage[T1]{fontenc}
\usepackage[utf8]{inputenc}
\usepackage{soul}
\usepackage{microtype}

\usepackage{inconsolata}
\usepackage{amsmath}
\usepackage{graphicx}
\usepackage{booktabs}
\usepackage[table]{xcolor}
\definecolor{darkgreen}{RGB}{0,128,0}
\definecolor{bestcolor}{RGB}{34, 139, 34}
\definecolor{lightgray}{RGB}{245, 245, 245}
\usepackage{latexsym}
\usepackage{tcolorbox}
\usepackage{fancyvrb}
\usepackage{listings}
\usepackage{colortbl}
\usepackage{multirow}

\usepackage{fancyhdr}
\title{Dicta-LM 3.0: Advancing The Frontier of Hebrew Sovereign LLMs}

\author{Shaltiel Shmidman\textsuperscript{1,†}, Avi Shmidman\textsuperscript{1,2,†}, 
Amir DN Cohen\textsuperscript{3,‡}
Moshe Koppel\textsuperscript{1,2,†} \\
\textsuperscript{1}DICTA / Jerusalem, Israel \\
\textsuperscript{2}Bar Ilan University / Ramat Gan, Israel \\ 
\textsuperscript{3}OriginAI / Ramat Gan, Israel \\ 
\texttt{\small \textsuperscript{†}\{shaltiel,avi,moishk\}@dicta.org.il} \\
\texttt{\small \textsuperscript{‡}{amirc@originai.co}}
}
 
\begin{document}
\maketitle
\thispagestyle{fancy}

\begin{abstract}

Open-weight LLMs have been released by frontier labs; however, sovereign Large Language Models (for languages other than English) remain low in supply yet high in demand. Training large language models (LLMs) for low-resource languages such as Hebrew poses unique challenges. In this paper, we introduce \texttt{Dicta-LM 3.0}: an open-weight collection of LLMs trained on substantially-sized corpora of Hebrew and English texts. The model is released in three sizes: 24B - adapted from the Mistral-Small-3.1 base model, 12B - adapted from the NVIDIA Nemotron Nano V2 model, and 1.7B - adapted from the Qwen3-1.7B base model. We are releasing multiple variants of each model, each with a native context length of 65k tokens; base model and chat model with tool-calling support. To rigorously evaluate our models, we introduce a new benchmark suite for evaluation of Hebrew chat-LLMs, covering a diverse set of tasks including Translation, Summarization, Winograd, Israeli Trivia, and Diacritization (\textit{nikud}). Our work not only addresses the intricacies of training LLMs in low-resource languages but also proposes a framework that can be leveraged for adapting other LLMs to various non-English languages, contributing to the broader field of multilingual NLP.

\end{abstract}

\section{Introduction}

The development of generative language models has significantly advanced natural language processing (NLP), enhancing the sophistication and contextual understanding in human-computer interactions \cite{openai2024gpt4technicalreport,geminiteam2024geminifamilyhighlycapable,claude2024model}. Leading state-of-the-art open-weight generative LLM models, such as Qwen and Olmo \cite{yang2025qwen3technicalreport,groeneveld2024olmo}, are trained on trillions of tokens, but only a very small percentage of that data represents low-resource languages such as Hebrew \cite{touvron2023llama2}. Consequently, these models significantly underperform in such languages.

Training large language models (LLMs) in low-resource languages presents unique challenges, stemming from limited data availability, complex morphological structures, and the lack of robust evaluation frameworks tailored to these languages. Hebrew, with its rich morphology and limited large-scale corpora, exemplifies these difficulties. 

In response to these challenges, we introduce \texttt{Dicta-LM 3.0}, a collection of generative language models specifically optimized for Hebrew. The models are trained with a native context length of 65k, and we release them in three sizes. The first is 24B parameters, built upon the Mistral-Small 3.1 model \cite{mistral_small_3_1}. The second is 12B parameter, built upon NVIDIA Nemotron Nano V2 \cite{nvidia2025nvidianemotronnano2}. The third is a 1.7B parameter built upon Qwen3 1.7B \cite{yang2025qwen3technicalreport}. All three models were continuously pre-trained on approximately 100 billion tokens of Hebrew, mixed together with 30B tokens of English data.

For each model size, we release multiple variants of the model: the base model and a chat model with tool-calling capabilities. 
We evaluate the base models on the Hebrew LLM Leaderboard \cite{shmidman2024adaptingllmshebrewunveiling}, a collection of Hebrew benchmarks evaluating base models using few-shot prompting \cite{brown2020languagemodelsfewshotlearners}. To address the evaluation gap for non-base Xomodels,
we introduce a new benchmark suite for assessing chat-style Hebrew language models. This suite includes tasks such as Translation, Summarization, Wingorad, Israeli Trivia, and Diacritization (\textit{nikud}). Our comprehensive evaluation demonstrates that \texttt{DictaLM3.0} achieve state-of-the-art performance on these tasks for its weight class.

The methodologies and evaluation frameworks presented in this work provide insights and potential pathways for adapting other LLMs to various non-English languages. This research contributes to the broader field of multilingual NLP by addressing the unique challenges posed by low-resource languages and offering scalable solutions for their integration into advanced language models.

\section{Pre-training Data}
\label{sec:pretraining-data}

Our pre-training corpus consisted of 75\% Hebrew and 25\% English data, detailed below. 

\subsection{Hebrew Data}

\subsubsection{Sources}

The Hebrew data consists of \textasciitilde100B tokens, collected from a wide range of sources including available open-source corpora, internal web scraping, Hebrew books scanned and digitized in-house, and partnerships with companies that graciously provided their internal data for training. The data can be broken down into the following categories:

\begin{itemize}

\item \textbf{Internet}: Approximately 65\% of the data. This category includes web crawls such as C4 \cite{2019t5}, OSCAR \cite{OrtizSuarezSagotRomary2019}, FineWeb2 \cite{penedo2025fineweb2pipelinescale}, Wikipedia, and others. We processed over 150TB of unprocessed crawl data when extracting this data. 

\item \textbf{Social Media}: Approximately 17\% of the data. This data includes various social media corpora such as Hebrew twitter posts and online Hebrew blogs. 

\item \textbf{News \& Legal}: Approximately 9\% of the data. This includes human transcriptions of TV and Radio, various news sites, Israeli parliament transcripts, and online legal resources. 

\item \textbf{Academia \& Literature}: Approximately 8.5\% of the data. This data includes academic papers, journals, books, and historical Hebrew texts (such as the Ben Yehuda project \cite{benyehuda_2024} and the Sefaria project \cite{sefaria_2024}). 

\item \textbf{Tagged Data}: Approximately 0.5\% of the data. This includes corpora tagged for various objectives such as NER, UD Dependencies \cite{tsarfaty2013unified,mcdonald2013universal,ZeldesHowellOrdanBenMoshe2022} and diacritics, as well as language resources such as dictionaries and thesauri. The data was reformatted into natural language explanatory text for the pretraining. 

\end{itemize}

\subsection{English Data}

\begin{figure}[h]
    \centering
    \includegraphics[width=0.7\textwidth]{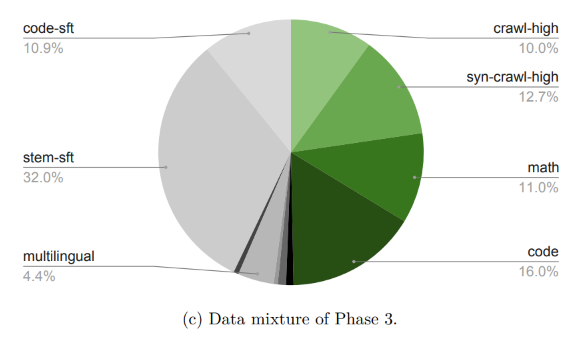}
    \caption{Data mixture used in Phase 3 of the Nemotron Nano V2 base model training. We used the same mixture in our CPT phase when mixing in the Nemotron-CC data.}
    \label{fig:nemo-data-mix}
\end{figure}

We include a 25\% mix of English data in our pretraining corpus, in order to mitigate catastrophic forgetting and to enable the model to apply the logic and reasoning present in the English training corpora to Hebrew input texts. For our training, we picked out three corpora:

\begin{itemize}

\item \textbf{Nemotron-CC}: This is the pretraining corpus used to train the NVIDIA Nemotron Nano V2 model \cite{nvidia2025nvidianemotronnano2}. We sample a mixture from the various subsets of this corpus, using the same parameters used in Phase 3 of the Nano V2 base model training - displayed in Figure \ref{fig:nemo-data-mix}.

\item \textbf{FineWeb-Edu} \cite{lozhkov2024fineweb-edu}: "1.3 trillion tokens of the finest educational data the web has to offer".

\item \textbf{SlimPajama} \cite{cerebras2023slimpajama}: "extensively deduplicated, multi-corpora, open-source dataset for training large language models."

\end{itemize}

For the continued pre-training of the Nemotron Nano V2 model, we \textit{only} used the Nemotron-CC data, whereas for the 24B model we split the English data 50/50 between Nemotron-CC and FineWeb-Edu. For the continued pre-training of the 1.7B model, we used the SlimPajama data mixed with 30\% data from the Llama Nemotron Post Training Dataset \cite{bercovich2025llamanemotronefficientreasoningmodels} in straight text format (not chat template), as the Nemotron-CC data had not been released yet. 

\section{Continuous Pre-training}
\label{sec:continue-pretraining}

Leading state-of-the-art generative LLM models are trained with trillions of tokens \cite{gemma_2024,llama3modelcard}, making their reproduction very expensive. Therefore, we initialize our model from an existing SOTA model and continue pretraining it, leveraging the vast amount of data the model was already trained on. We chose to initialize from the following models, continuing to train each one separately:

\begin{itemize} 

\item \textbf{Mistral-Small-3.1-24B-Base-2503}\footnote{\url{https://huggingface.co/mistralai/Mistral-Small-3.1-24B-Base-2503}} \cite{mistral_small_3_1} 

\item \textbf{NVIDIA-Nemotron-Nano-12B-v2-Base}\footnote{\url{https://huggingface.co/nvidia/NVIDIA-Nemotron-Nano-12B-v2-Base}} \cite{nvidia2025nvidianemotronnano2} 

\item \textbf{Qwen3-1.7B-Base}\footnote{\url{https://huggingface.co/Qwen/Qwen3-1.7B-Base}} \cite{yang2025qwen3technicalreport} 

\end{itemize}

Byte-pair encoding (BPE) tokenizers in foundation models often inadequately cover low-resource languages due to their scarcity in training data, leading to common words being split into many sub-tokens. This reduces compression, and therefore when choosing baseline models we had to ensure that the models had a good ratio, making them more practical for Hebrew.

We conducted the continuous pre-training on a compute cluster of 80 H200 141GB GPUs on NVIDIA DGX Cloud Lepton \cite{anand_soman_2025_dgx_cloud_lepton}. All training was done using the NVIDIA NeMo Framework \cite{Harper_NeMo_a_toolkit}, a scalable generative AI framework built for researchers and developers working on large language models, optimized for large-scale model training on NVIDIA hardware. Lepton’s unified AI platform made it simple to scale distributed training across H200 GPUs and manage experiments through NeMo-Run, cutting iteration time and infrastructure overhead.

We split the continuous pre-training into two phases. In the first stage we iterated over our entire dataset, with a context window of 4,096 tokens. In the second stage we re-iterated over 20\% of the dataset, with an extended context window of 65k tokens.

\subsection{Phase 1}

\begin{figure}[h]
    \centering
    \includegraphics[width=.95\linewidth]{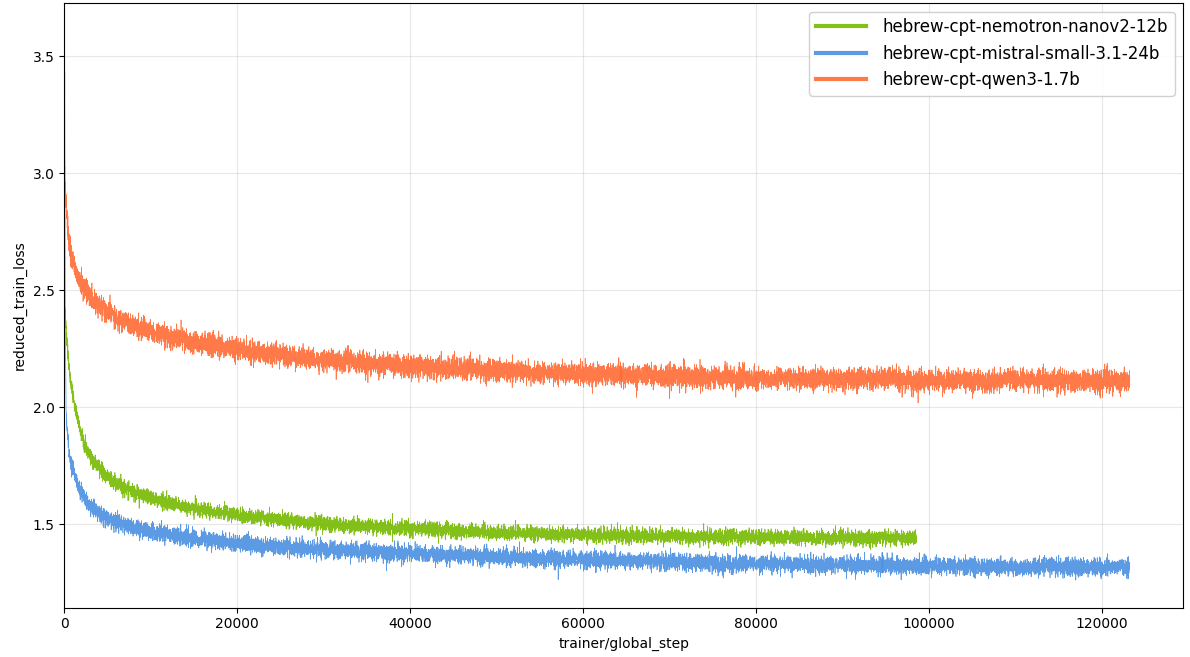}
    \caption{This graph depicts the loss value during the continuous pre-training stage; the 12B model had a larger batch size, which required fewer steps.}
    \label{fig:train-loss}
\end{figure}

In the first phase we iterated over our entire dataset - 75\% Hebrew and 25\% English - summing to a total of 130B tokens. More details about the training are listed in Table \ref{tab:training-hyp}; the training loss graph is shown in Figure \ref{fig:train-loss}. 

\begin{table}[h]
\centering
\begin{tabular}{llll}
\hline
\textbf{Hyperparameter} & \textbf{24B} & \textbf{12B} & \textbf{1.7B} \\
\hline
Global batch size (GBS)\footnotemark & 256 & 320 & 256 \\
Micro batch size (MBS) & 4 & 1 & 8\\
Tensor Parallel (TP) & 4 & 1 & 1\\
Learning rate (LR) & 5e-6 & 5e-6 & 1e-5 \\
\hline
Sequence length & \multicolumn{3}{c}{4096} \\
Warmup steps & \multicolumn{3}{c}{2000} \\
Optimizer & \multicolumn{3}{c}{Distributed AdamW} \\
Schedule & \multicolumn{3}{c}{Cosine annealing} \\
Adam betas & \multicolumn{3}{c}{$\beta_1 = 0.9,\ \beta_2 = 0.95$} \\
Clip grad & \multicolumn{3}{c}{1.0} \\
Weight decay & \multicolumn{3}{c}{0.1} \\
Total tokens & \multicolumn{3}{c}{\(\sim 130\text{B}\)} \\
\hline
\end{tabular}
\caption{Training hyperparameters for phase 1 of the continuous pre-training.}
\label{tab:training-hyp}
\end{table}

\footnotetext{GBS differs because training capacity increased from 64 to 80 devices, requiring GBS to be divisible by 80.}

\subsection{Phase 2}

In the second phase we extended the model's context length to 65k tokens. For this phase, we split our data into documents with >6144 tokens and <6144 tokens. We then continued to train the model with the extended context length, sampling 75\% of the documents from the longer context pool, and 25\% of the documents from the shorter context pool. We maintained the Hebrew-English ratio of 75/25, training for a total of 18B tokens. The details for this phase of the training are detailed in Table \ref{tab:training-hyp-long}.

\begin{table}[h]
\centering
\begin{tabular}{llll}
\hline
\textbf{Hyperparameter} & \textbf{24B} & \textbf{12B} & \textbf{1.7B} \\
\hline
Learning rate (LR) & 1e-6 & 1e-7 & 6e-7 \\
Tensor Parallel (TP) & 4 & 1 & 1 \\
Context Parallel (CP) & 5 & 16 & 8 \\
Micro batch size (MBS) & 1 & 1 & 2\\
Global batch size (GBS) & 80 & 80 & 16 \\
Sequence length & 65{,}280 & 65{,}280 & 62{,}080 \\
Warmup steps & 200 & 200 & 1\% \\
\hline
\end{tabular}
\caption{Training hyperparameters for phase 2 of the continuous-pretraining.}
\label{tab:training-hyp-long}
\end{table}

\section{Base Model Evaluation}
\label{sec:base-model-eval}

\begin{figure}[h]
    \centering
    \includegraphics[width=1\textwidth]{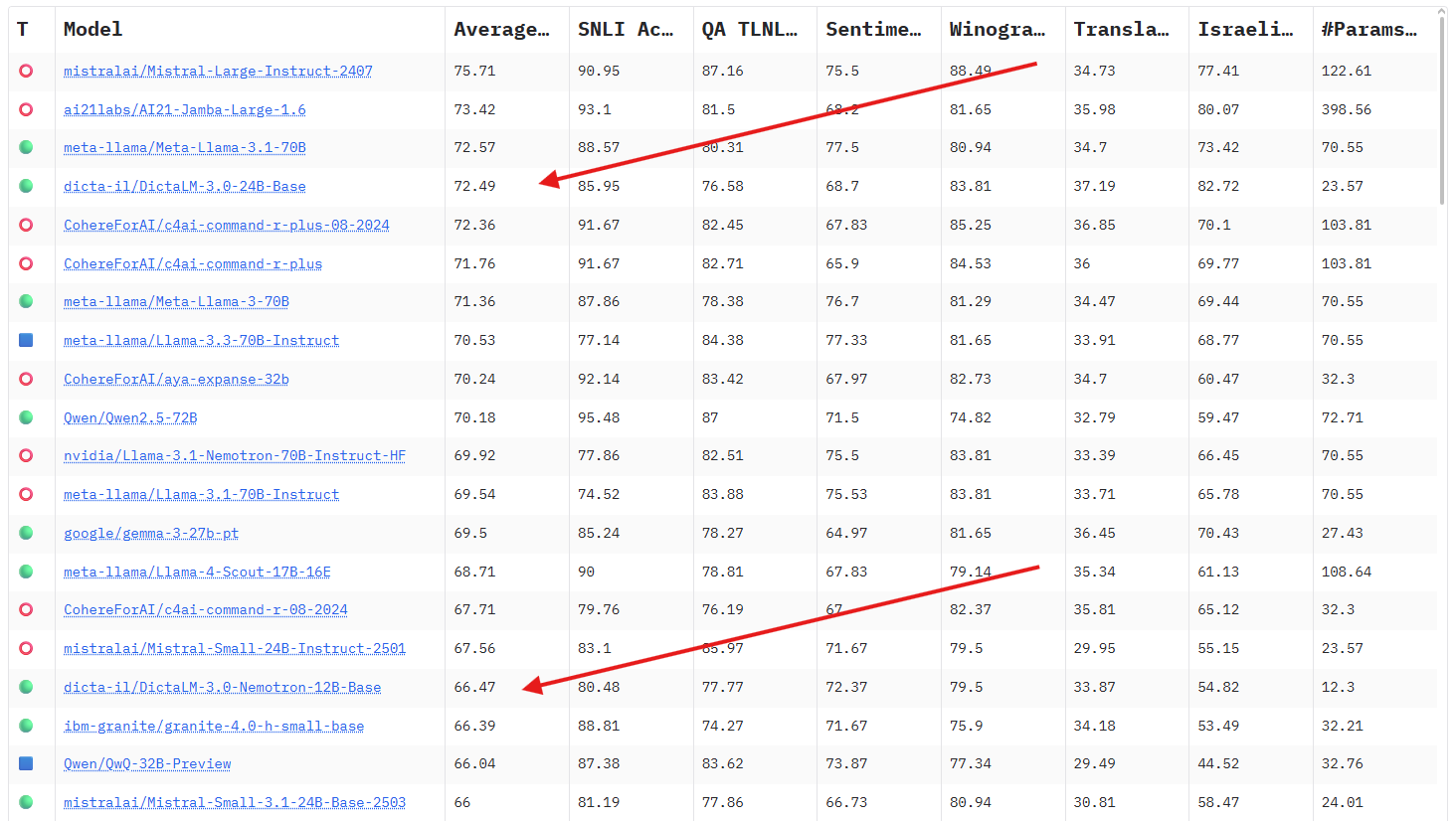}
    \caption{Up to date screenshot from the Hebrew LLM-Leaderboard, as of the publication date}
    \label{fig:leaderboard}
\end{figure}

We evaluated the base models on Hebrew benchmarks from the Hebrew LLM-Leaderboard \cite{hebrew_llm_leaderboard_hf_space,shmidman2024adaptingllmshebrewunveiling}. We observe significant gains relative to the base models we initialized from, in the 24B, 12B, and 1.7B models. A full screenshot from the leaderboard including the 24B and 12B new base models is displayed in Figure \ref{fig:leaderboard}. A before/after comparison of the three models can be found in Table \ref{tab:before-after-base-eval}. Notably, the 24B base model matches or exceeds the performance of models over four times larger, while also attaining the top score in the Israeli Trivia category, highlighting the effectiveness of sovereign LLMs.

We also evaluated the model on English benchmarks, in order to make sure the model retained the knowledge and abilities it was initially trained on. We evaluated the model on 3 popular benchmarks: Commonsense QA \cite{talmor-etal-2019-commonsenseqa}, WinoGrande \cite{sakaguchi2019winograndeadversarialwinogradschema} and Arc-Challenge \cite{clark2018thinksolvedquestionanswering}. The evaluation was done using the \texttt{lighteval} library \cite{lighteval}. The exact command that we ran can be found in Appendix \ref{appendix:leval-eng-command}. As can be seen in the table, we managed to retain >98\% of the English capabilities of the original models.

Both base models are available for use on HuggingFace with a permissive license.

\begin{table}[h]
\centering
\small
\begin{tabular}{@{}lccccccc@{}}
\toprule
\textbf{Model} & \textbf{SNLI} & \textbf{QA} & \textbf{Sentiment} & \textbf{Winograd} & \textbf{Translation} & \textbf{IL-Facts} & \textbf{Average} \\
\midrule
\multicolumn{8}{@{}l@{}}{\textit{24B Parameter Models}} \\[2pt]
Mistral-Small-3.1-24B & 81.2 & \textbf{77.9} & 66.7 & 80.9 & 30.8 & 58.5 & 66.0 \\
\rowcolor{blue!8}
DictaLM-3.0-24B & \textbf{86.0} & 76.6 & \textbf{68.7} & \textbf{83.8} & \textbf{37.2} & \textbf{82.7} & \textbf{72.5} \\
\textit{Improvement} & \textcolor{darkgreen}{+4.8} & \textcolor{red}{-1.3} & \textcolor{darkgreen}{+2.0} & \textcolor{darkgreen}{+2.9} & \textcolor{darkgreen}{+6.4} & \textcolor{darkgreen}{+\textbf{24.2}} & \textcolor{darkgreen}{+\textbf{6.5}} \\[6pt]
\midrule
\multicolumn{8}{@{}l@{}}{\textit{12B Parameter Models}} \\[2pt]
Nemotron-Nano-12B-v2 & 71.4 & 66.1 & 64.5 & 64.0 & 21.5 & 34.9 & 53.7 \\
\rowcolor{blue!8}
DictaLM-3.0-Nemotron-12B & \textbf{80.5} & \textbf{77.8} & \textbf{72.4} & \textbf{79.5} & \textbf{33.9} & \textbf{54.8} & \textbf{66.5} \\
\textit{Improvement} & \textcolor{darkgreen}{+9.1} & \textcolor{darkgreen}{+11.7} & \textcolor{darkgreen}{+7.9} & \textcolor{darkgreen}{+15.5} & \textcolor{darkgreen}{+12.4} & \textcolor{darkgreen}{+\textbf{19.9}} & \textcolor{darkgreen}{+\textbf{12.8}} \\
\midrule
\multicolumn{8}{@{}l@{}}{\textit{1.7B Parameter Models}} \\[2pt]
Qwen3-1.7B-Base & 46 & 61.4 & 53 & 54.3 & 20.1 & 24.9 & 43.3 \\
\rowcolor{blue!8}
DictaLM-3.0-1.7B & \textbf{67.1} & \textbf{62.8} & \textbf{61.9} & \textbf{57.2} & \textbf{29.2} & \textbf{30.6} & \textbf{51.5} \\
\textit{Improvement} & \textcolor{darkgreen}{+21.1} & \textcolor{darkgreen}{+1.4} & \textcolor{darkgreen}{+8.9} & \textcolor{darkgreen}{+2.9} & \textcolor{darkgreen}{+9.1} & \textcolor{darkgreen}{+5.7} & \textcolor{darkgreen}{+\textbf{8.2}} \\
\bottomrule
\end{tabular}
\caption{Performance comparison of the base models on Hebrew benchmarks before and after continuous pre-training}
\label{tab:before-after-base-eval}
\end{table}

\begin{table}[h]
\centering
\small
\begin{tabular}{@{}lcccc@{}}
\toprule
\textbf{Model} & \textbf{CommonSense} & \textbf{WinoGrande} & \textbf{Arc-Challenge} & \textbf{Avg.} \\
\midrule
\multicolumn{5}{@{}l@{}}{\textit{24B Parameter Models}} \\[2pt]
Mistral-Small-3.1-24B & 83.2 & 80.0 & 65.1 & 76.1 \\
\rowcolor{blue!8}
DictaLM-3.0-24B & 82.8 & 78.2 & 63.4 & 74.8 \\
\textit{Difference} & \textcolor{red}{-0.4} & \textcolor{red}{-1.8} & \textcolor{red}{-1.7} & \textcolor{red}{-1.3} \\[6pt]
\midrule
\multicolumn{5}{@{}l@{}}{\textit{12B Parameter Models}} \\[2pt]
Nemotron-Nano-12B-v2 & 79.4 & 75.7 & 64.7 & 73.3 \\
\rowcolor{blue!8}
DictaLM-3.0-Nemotron-12B & \textbf{80.7} & \textbf{76.2} & 64.4 & \textbf{73.8} \\
\textit{Difference} & \textcolor{darkgreen}{+1.2} & \textcolor{darkgreen}{+0.5} & \textcolor{red}{-0.3} & \textcolor{darkgreen}{+0.5} \\
\midrule
\multicolumn{5}{@{}l@{}}{\textit{1.7B Parameter Models}} \\[2pt]
Qwen3-1.7B-Base & 74.4 & 61.5 & 38.8 & 58.2 \\
\rowcolor{blue!8}
DictaLM-3.0-1.7B & 70.2 & 59.8 & \textbf{50.6} & \textbf{60.2} \\
\textit{Difference} & \textcolor{red}{-4.2} & \textcolor{red}{-1.7} & \textcolor{darkgreen}{+11.8} & \textcolor{darkgreen}{+2.0} \\
\bottomrule
\end{tabular}
\caption{Performance comparison of the base models on English benchmarks}
\label{tab:english-base-eval}
\end{table}

\section{Post-training}

Following the creation of our \texttt{DictaLM-3.0} base models via continuous pre-training, we produced variants of the models - chat models - by post-training the models on conversational data.

As part of our post-training pipeline, we produce a chat model - either instruct (GPT-3.5 style), or thinking (o1 style).  

\subsection{Supervised Fine-Tuning (SFT)}

\subsubsection{SFT Data}
\label{sec:sft-data}

For the first stage of post-training, we curated two conversational datasets for supervised fine-tuning. The first is the ‘instruct’ dataset, intended for regular chat conversations with direct answers. The second is the ‘thinking’ dataset, which adds a designated reasoning section before each assistant response. 

For this stage, we aim for a 50/50 mix of English and Hebrew data. For most datasets, we obtain the Hebrew versions by translating them from their corresponding English versions with the \texttt{DictaLM-3.0} base model, using few-shot prompting. We also apply a filter to remove conversations that cannot be meaningfully translated (e.g., translation-request dialogues that would collapse into identical Hebrew text and introduce noise).

We gather our data from various open-source datasets. Following the findings by \citet{shmidman2025learningreasontrainingllms}, we use \texttt{gpt-oss-120b} \cite{openai2025gptoss120bgptoss20bmodel} to regenerate the responses of the reasoning datasets.
In addition, all datasets undergo post-processing to normalize formats and filter conversations through a rule-based system (e.g., removing refusals).

Below is an itemized list of the datasets in our final post-training SFT corpus. For datasets used for both thinking and instruct, the instruct version is derived by simply removing the reasoning section.

\begin{itemize}
    \item Hermes 3 Dataset \cite{Hermes3Dataset}: We use 200k sampled conversations and include them along with their Hebrew translations. \textit{Instruct only}.

    \item Hermes 3 - Augmented Subset: We sample 70k conversations that include a system message, take the first user prompt, and generate multi-turn dialogues using \texttt{gpt-oss-120b} (the model generating both stages separately). We then translate all conversations to Hebrew and interleave the languages, often adding small augmentations (e.g., adding ‘Answer only in Hebrew’ to the system prompt, adding ‘answer in English,’ or simple context-switching to teach the model to follow the user’s language). \textit{Thinking only}.

    \item Math \cite{shmidman2025learningreasontrainingllms}: We use  250k sampled conversations from the Nemotron Post Training Dataset \cite{NemotronPostTrainingDatasetV1,bercovich2025llamanemotronefficientreasoningmodels}, with a verified answered generated via \texttt{gpt-oss-120b}. We include them along with their Hebrew translations. \textit{Instruct \& Thinking}. 

    \item rStarCoder \cite{liu2025rstarcoderscalingcompetitivecode}: In this work we did not focus on code-generation abilities, since code is written in English and our emphasis is on Hebrew. Nevertheless, we include a small sample of code-related conversations, as previous work shows this can improve overall model capabilities. We use 30k sampled conversations and include them along with their Hebrew translations. \textit{Instruct \& Thinking}. 

    \item Everyday \& System conversations (SmolTalk v2 \cite{bakouch2025smollm3}): This dataset includes 30k examples of both everyday conversations and dialogues with a system prompt (teaching the model to adjust behavior based on the system message). The thinking subsets were regenerated using \texttt{gpt-oss-120b}. We include them along with their Hebrew translations. \textit{Instruct \& Thinking}.

    \item Tulu3 SFT Personas IF \cite{lambert2025tulu3pushingfrontiers}: We do not include Hebrew translations for this dataset, as it contains very specific style constraints that do not translate well (e.g., requiring responses in all caps). \textit{Instruct only}.

    \item CCMatrix Translation \cite{schwenk2020ccmatrixminingbillionshighquality}: We sample 300k English–Hebrew pairs and convert them into conversational data using \textasciitilde30 different templates. \textit{Instruct only}.

    \item Agent-Ark/Toucan-1.5M \cite{xu2025toucan}: We sample 65k conversations from the SFT subset for the instruct data, and 50k conversations from the OSS subset for the thinking data. We include all conversations along with their Hebrew translations, with post-processing to ensure that all tool-calling JSON outputs are valid and match the expected schema.  \textit{Instruct \& Thinking}.

    \item Identity: We auto-generated 3k conversations that include information about the model's identity (e.g., \texttt{'Who are you?'} -> \texttt{'I am Dicta-LM..'}). \textit{Instruct \& Thinking}.

    \item Others: We include about 100k other conversations that were synthesized directly by us. This includes mostly taking supervised datasets and converting them to conversational data using templates. For instance, given an annotated Hebrew sentence with morphological tagging, a conversation is generated in which a request is made regarding the morphological characteristics of a word with the sentence, and the response provides the relevant answer. 
    
\end{itemize}

The final combined dataset for instruct consisted of \textasciitilde2B tokens, from 1.5M conversations. The final combined dataset for reasoning consisted of \textasciitilde3.2B tokens, from 725k conversations. 

For the chat template, we drew inspiration from various existing models. For the message delimiter tokens we followed the convention of the Qwen3 models \cite{yang2025qwen3technicalreport}, for tool calling we followed the Hermes convention \cite{teknium2024hermes3technicalreport}, and for reasoning we followed the DeepSeek-R1 convention \cite{deepseekai2025deepseekr1incentivizingreasoningcapability}. We allocated the following special tokens in the tokenizer: \texttt{<|im\_start|>}, \texttt{<|im\_end|>}, \texttt{<tool\_response>}, \texttt{</tool\_response>}, \texttt{<tool\_call>}, \texttt{</tool\_call>}, \texttt{<think>}, \texttt{</think>}.

\subsubsection{Training}

We conducted the supervised fine-tuning on the same compute cluster on NVIDIA DGX Cloud Lepton, using NVIDIA NeMo Framework \cite{Harper_NeMo_a_toolkit}.

We packed all the data into sequences of 65k tokens using the first-fit-decreasing method. For this phase of training, we computed the loss only on the completion tokens (the assistant responses). In addition, to maintain the knowledge gained during the pre-training phase, we mix in 10\% pre-training data.

We first train the model for 10 epochs on the instruct data, and then we subsequently train the model for 5 epochs on the reasoning data. The hyperparameters can be found in Table \ref{tab:training-hyp-sft}.

\begin{table}[h]
\centering
\begin{tabular}{llll}
\hline
\textbf{Hyperparameter} & \textbf{24B} & \textbf{12B} & \textbf{1.7B} \\
\hline
Learning rate (LR) & 5e-6 & 1e-6 & 1e-5\\
Tensor Parallel (TP) & 4 & 1 & 1\\
Context Parallel (CP) & 5 & 16 & 4 \\
Sequence length & 65{,}280 & 65{,}280 & 62{,}080 \\
Global batch size (GBS) & 40 & 48\footnotemark & 48 \\
\hline
Micro batch size (MBS) & \multicolumn{3}{c}{1} \\
Warmup ratio & \multicolumn{3}{c}{0.03} \\
\hline
\end{tabular}
\caption{Training hyperparameters for the SFT stage.}
\label{tab:training-hyp-sft}
\end{table}

\footnotetext{Trained on 96 GPUs; global batch size chosen to remain divisible by 6.}

\subsection{Direct Preference Optimization (DPO)}

The next stage in our model refinement process was Direct Preference Optimization (DPO)  \cite{rafailov2023directpreferenceoptimizationlanguage}. DPO focuses on optimizing the model based on user preferences and feedback, enhancing its ability to generate responses that are both accurate and aligned with user expectations. We focused DPO on instruct variants due to the limited availability of diverse models employing the \texttt{gpt-oss} reasoning style, which is necessary for the comparative preference DPO requires.

\subsubsection{DPO Data}

We started off with the \texttt{llama-3.1-tulu-3-70b-preference-mixture} dataset \cite{lambert2025tulu3pushingfrontiers}, which contains close to 300k preference pairs. We start by filtering out any pair that wasn't in English, and then sampling 150k examples from the dataset. As with the SFT data, we translated the 150k examples to Hebrew using the \texttt{Dicta-LM 3.0} 24B base model, so that our final dataset consists of a 50/50 mix of English and Hebrew. 

As shown by \cite{lambert2025tulu3pushingfrontiers}, it is important to mix on-policy examples as well. Therefore, we sampled 15\% of our final dataset, and generated a response by passing the prompt to our 24B-Instruct model. We then used GPT-4o \cite{openai2024gpt4ocard} to compare our on-policy generated response with the original preferred response, to determine whether our response is preferred.  

In addition, we mix in a few thousand examples of the following two types of behavioral preference data:

\begin{itemize}
    \item We compiled the identity examples from our SFT corpus, and then used an LLM to generate negative examples (e.g., \texttt{You are ChatGPT, ...}) reinforcing the Dicta-LM identity in the model.

    \item We observed an interesting behavior during testing: after receiving a large tool response (such as a web-search result) in a language different from the original prompt, the model would continue responding in that new language. To address this, we generated synthetic preference pairs by sampling from the model, appending the instruction “Make sure your final response is in Hebrew” (or English) after the tool response, and then removing that instruction from the final preference pair. 
\end{itemize}

As these two types of preference data were generated by artifically creating negative examples, we also generated similar preference pairs with the thinking section using our thinking model. 

The final dataset consists of 260k examples. 

\subsubsection{Training}

We conducted the supervised fine-tuning on the same compute cluster on NVIDIA DGX Cloud Lepton, using the NVIDIA \textbf{NeMo-RL} Training Framework \cite{nemo-rl}. The NeMo-RL library integrated directly with Lepton, providing us with a seemless experience for scaling up post-training. 

We ran the DPO training using full parameter training (as opposed to PEFT), utilizing Megatron-LM as the backend \cite{megatron-lm} achieving a near 2x speedup. For the instruct models, we run a full epoch on top of the SFT model. For the thinking models, we run 30 training steps on the two behavioral subsets on top of the GRPO model (see below). The full hyperparameters are listed in Table \ref{tab:training-hyp-dpo}.

\begin{table}[h]
\centering
\begin{tabular}{ll}
\hline
\textbf{Hyperparameter} & \textbf{Value} \\
\hline
Max total sequence length & 4{,}096 \\
Global batch size (GBS) & 144 \\
Learning rate (LR) & 5e{-}7 \\
Minimum LR & 5e{-}8 \\
LR decay style & cosine \\
LR warmup iterations & 50 \\
Tensor Parallel (TP) & 4 \\
\hline
\end{tabular}
\caption{Training hyperparameters for DPO training.}
\label{tab:training-hyp-dpo}
\end{table}

\subsection{Group Relative Policy Optimization (GRPO)}

The next stage in our model refinement process was Group Relative Policy Optimization (GRPO) \cite{shao2024deepseekmathpushinglimitsmathematical}. GRPO directly trains the policy using grouped comparisons, improving the model’s reasoning quality and overall response reliability. We applied GRPO to our thinking-style variants, as recent work has shown that GRPO often yields the largest improvements in reasoning and alignment quality.

\subsubsection{GRPO Data}

A key part of GRPO training is being able to calculate a reward for a generated response to a given prompt. In order to be able to accomplish that, we carefully curate a dataset, where we include metadata with each prompt allowing us to verify the correctness of a given answer and assign a numerical reward (between 0 and 1). Our dataset comprises of the following tasks:

\begin{itemize}
    \item Instruction-Following \cite{zhou2023instructionfollowingevaluationlargelanguage}: This task defines a set of instructions which are appended to other prompts, defining how the response should look (e.g., no capital letters). Each prompt includes a verifier function which can be run on the final response to assess whether the constraint was met. We use the \texttt{allenai/RLVR-IFeval} dataset for English examples. For Hebrew, we manually curate a set of 30 Hebrew-specific constraints, and then sample prompts from our SFT data and randomly append one of our Hebrew-specific constraints to each prompt. The final score is either 0 (constraint wasn't followed) or 1 (constraint was followed). 

    \item MATH \cite{hendrycksmath2021} \& GSM8K \cite{cobbe2021gsm8k}: This dataset includes math questions with their correct answer included in the metadata. We can then verify the answer using the HF Math-Verify library \cite{Kydlicek_Math-Verify_Math_Verification}. We take the data as-is from \texttt{allenai/RLVR-MATH} and \texttt{allenai/RLVR-GSM}, and include them along with their Hebrew translations. The final score is either 0 (incorrect) or 1 (correct). 

    \item Nikud (Diacriticization): This task is Hebrew-specific, where we provide the model with a sentence in Hebrew and expect the output with diacritics. This is an increasingly difficult task for LLMs to accomplish, as they both have to disambiguate the different meanings of each word, and also break down each word to the indivdiual letters as the diacritics are separate tokens. We use our in-house curated corpus of sentences diacritized by linguistic experts, providing us with a gold corpus to verify against. The final score is the percent of words correctly diacritized (between 0 and 1). 

    \item Universal Dependencies (UD) Parsing: This task evaluates syntactic analysis by providing a sentence and expecting a full Universal Dependencies parse, including part-of-speech tags, dependency heads, and dependency relations. The model must correctly analyze the sentence and produce a correct by-word breakdown of the sentence. This is a very non-trivial task that even frontier-LLMs struggle with. We use the data from the Hebrew Treebank \cite{ZeldesHowellOrdanBenMoshe2022,tsarfaty2013unified,mcdonald2013universal} as our gold standard (selecting only sentences which were not included in the earlier training phases). The final score is the percent of correct labels produced (between 0 and 1). 
    
\end{itemize}

The final dataset consists of 46k examples. 

\subsubsection{Training}

As with DPO training, we conducted the supervised fine-tuning on the same compute cluster on NVIDIA DGX Cloud Lepton, using the NVIDIA NeMo-RL Training Framework.

For this train we use the Megatron-LM backend for the training, and the vLLM backend \cite{kwon2023efficient} for generation. We detail the full training parameters in Table \ref{tab:training-hyp-grpo}. 

\begin{figure}[h]
    \centering
    \includegraphics[width=0.7\textwidth]{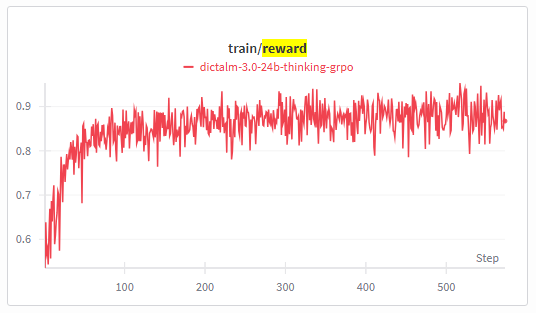}
    \caption{Reward graph when running GRPO training on the 24B thinking variant.}
    \label{fig:grpo-reward}
\end{figure}

We ran GRPO training on the \texttt{Dicta-LM 3.0} 24B and 1.7B thinking variants, with the model quickly learning. The reward graph for the 24B can be seen in Figure \ref{fig:grpo-reward}, and as can be seen the model quickly improves, and begins to plateau after 250 steps. 

\begin{table}[h]
\centering
\begin{tabular}{ll}
\hline
\textbf{Hyperparameter} & \textbf{Value} \\
\hline
Micro batch size & 2 \\
Global batch size & 480 \\
Tensor Parallel (TP) & 4 \\
Max total sequence length & 8{,}192 \\
Num prompts per step & 64 \\
Num generations per prompt & 32 \\
KL penalty & 0.01 \\
KL type & k3 \\
Learning rate (LR) & 3e{-}7 \\
\hline
\end{tabular}
\caption{Training hyperparameters for the GRPO train.}
\label{tab:training-hyp-grpo}
\end{table}

\section{Chat Model Evaluation}

\subsection{Hebrew Chat Evaluation}

For Hebrew chat evaluation, we curated a set of five tasks to evaluate the performance of LLMs on Hebrew. The evaluation here differs from base model evaluation, where for the base model we provide few-shot examples, and here the model is provided with a 0-shot prompt. The tasks are also available in the form of a public leaderboard.\footnote{\url{https://huggingface.co/spaces/hebrew-llm-leaderboard/chat-leaderboard}} The five evaluation tasks are as follows:

\begin{itemize}
    \item \textbf{Summarization}: Evaluating the model's ability to summarize a document. The test corpus is comprised of 1,000 documents of various styles (news, wiki, etc.). We compare the test model's generated summary with a summary generated via Gemini-2.5-Pro, and check the win rate. If Gemini-2.5-Pro is rated higher, we assign a score of 0; if the test model's summary is better, we assign a score of 1. If the two are equally good, we assign a score of 0.5. We use GPT4o as the LLM as a Judge to determine the score here.

    \item \textbf{Translation}: The test corpus consists of a random set of  English paragraphs, 20-40 words in length. For each one, we ask the test model to generate a translation to Hebrew, and we compare to the translation generated from Gemini-2.5-Pro, utilizing the same approach as detailed above with regard to summarization. 

    \item \textbf{Israeli Trivia}: The test corpus consists of 300 trivia questions regarding Israel-related facts. We extend immense gratitude to Avraham Elitzur for helping us curate this dataset. We provide multiple-choice answers in the prompt and compute the accuracy score using exact string matching. 

    \item \textbf{Winograd}: We use "A Translation of the Winograd Schema Challenge to Hebrew" \cite{shwartz2021wsc}. Models must resolve ambiguous pronouns using commonsense reasoning. We provide the ambiguous sentence and the question, with two choices for the model to choose from. 

    \item \textbf{Nikud (Diacritization)}: We use a test corpus of sentences that were manually diacritized by expert linguists (ensuring that none were present in any phase of the training). Each sentence is scored according to the number of words were correctly diacritized. The final score is calculated as the macro average of the accuracies across the sentences in the corpus. 

\end{itemize}

We present the results of our models, comparing them to other models of similar sizes in Table \ref{tab:dictalm-benchmarks}. As can be seen, the 24B model performs extremely well on language-specific tasks, which usually require a more subtle understanding of the language. In addition, the model also performs on par / better than models >30\% larger than it on tasks which require more extensive world knowledge. The Nemotron-12B performs very well competitive with gemma-3-12b-it, yet with the major advantage that the Nemotron model is based on a hybrid-SSM architecture, allowing it to scale across longer context lengths with higher throughput and a lower memory footprint. The 1.7B model showcases remarkable performance for its size significantly outperforming other models in that size range. 

\begin{table}[h]
\centering
\small
\begin{tabular}{lccccc}
\toprule
\textbf{Model} & \textbf{Summarization} & \textbf{Translation} & \textbf{Winogrande} & \textbf{Trivia} & \textbf{Nikud} \\
\midrule
\textbf{DictaLM-3.0-24B-Thinking}       & \textbf{56.86} & \textbf{30.09} & 78.06 & \textbf{60.13} & \textbf{86.86} \\
gemma3-27B-it                       & 44.54 & 26.73 & \textbf{79.86} & 45.51 & 60.21 \\
aya-expanse-32B                     & 29.46 & 17.10 & 80.58 & 53.82 & 45.40 \\
Llama-3.3-70B-Instruct              & 37.83 & 19.31 & 83.45 & \textbf{60.13} &  4.05 \\
\midrule
\multicolumn{6}{c}{\textbf{Smaller models (\textasciitilde12B)}} \\
\midrule
\textbf{DictaLM-3.0-Nemotron-12B-Instruct}             & 33.27 & 13.50 & 73.74 & \textbf{45.18} & \textbf{76.12} \\
gemma-3-12b-it                                         & \textbf{39.48} & \textbf{16.50} & \textbf{75.90} & 44.85 & 51.78 \\
Qwen3-14B (think)                                             & 15.83 &  0.90 & 73.38 & 41.86 &  4.73 \\
\midrule
\multicolumn{6}{c}{\textbf{Tiny models (\textasciitilde1.7B)}} \\
\midrule
DictaLM-3.0-1.7B-Instruct                              & 9.72 & 2.16 & \textbf{58.2} & 30.21 & \textbf{52.76} \\ 
\textbf{DictaLM-3.0-1.7B-Thinking}                              & \textbf{10.22} & \textbf{2.51} & 55.76 & \textbf{31.23} & 47.2 \\ 
gemma-3-1b-it                              & 0.35 & 0.15 & 47.84 & 26.58 & 3.44 \\ 
Qwen3-1.7B (think)                             & 0.4 & 0 & 51.08 & 21.59 & 2.93 \\ 

\bottomrule
\end{tabular}
\caption{Evaluation scores for DictaLM-3.0 variants and baselines (higher is better).}
\label{tab:dictalm-benchmarks}
\end{table}

\subsection{English Chat Evaluation}

For the English chat evaluation, we evaluate using the Olmes library \cite{allenai_olmes} on the same suite of tasks as Olmo3 \cite{Olmo3_2025}\footnote{We don't evaluate on the coding tasks, as we specifically did not focus on code generation abilities (see Section \ref{sec:sft-data}). In addition, we didn't evaluate on the following tasks since the datasets used in olmes aren't public as of the publication date: AIME 2024, AIME 2025, IFBench}. We present the results for the models in Tables \ref{tab:res-midsize-models} and \ref{tab:res-12b-models}. \texttt{DictaLM 3.0} 24B showcases very strong results across the board on the English benchmarks, specifically with regard to instruction following and chat, making it an ideal choice for downstream use. \texttt{DictaLM 3.0} 12B shows strong knowledge retention, and performing competitively on the mathematical benchmarks. 

In Table \ref{tab:thinking-efficiency} we compare the behavior of the instruct and thinking variants of the 24B model, analyzing the efficiency of the model versus the resulting accuracy gains. Similar to the findings of \citet{shmidman2025learningreasontrainingllms}, we observe that the thinking model delivers notable accuracy improvements without becoming overly verbose. In several cases, it even achieves a significant boost in accuracy while generating \textit{fewer} tokens. We believe this occurs because the reasoning content is more compact and efficient than standard output, enabling the model to produce high-quality answers without incurring substantial inference costs. 

\begin{table}[h]
\small
\centering
\begin{tabular}{@{}llccc@{}}
\toprule
\textbf{Category} & \textbf{Benchmark} & \textbf{DictaLM 3.0 24B-Think} & \textbf{Gemma 3 27B} & \textbf{Mistral Small 3.1} \\
\midrule
\multirow{2}{*}{Math} 
    & MATH          & 86.41 & \textbf{87.40} & 67.93 \\
    & OMEGA         & \textbf{28.38} & 24.00 & 13.76 \\
\midrule
\multirow{3}{*}{Reasoning} 
    & BigBenchHard  & 73.00 & \textbf{82.40} & 71.91 \\
    & ZebraLogic    & \textbf{47.80} & 24.80 & 23.10 \\
    & AGI Eval (EN) & \textbf{82.93} & 76.90 & 75.87 \\
\midrule
Inst. Following & IFEval & \textbf{88.17} & 85.40 & 79.48 \\
\midrule
\multirow{3}{*}{Knowledge} 
    & MMLU          & \textbf{85.93} & 74.60 & 83.11 \\
    & PopQA         & \textbf{30.59} & 30.20 & 29.70 \\
    & GPQA          & \textbf{55.13} & 45.00 & 44.87 \\
\midrule
Chat & AlpacaEval 2 LC & \textbf{74.11} & 65.50 & 41.22 \\
\bottomrule
\end{tabular}
\caption{Performance comparison of mid-size models (24B--27B parameters). DictaLM-3.0-24B-Thinking achieves the strongest results on instruction following and chat, while Gemma 3 27B leads on several reasoning tasks.}
\label{tab:res-midsize-models}
\end{table}

\begin{table}[h]
\centering
\small
\begin{tabular}{@{}ll cc@{}}
\toprule
\textbf{Category} & \textbf{Benchmark} & \textbf{Gemma 3 12B} & \textbf{DictaLM 3.0 12B-Inst} \\
\midrule
\multirow{2}{*}{Math} 
    & MATH              & \textbf{84.25} & 74.99 \\
    & OMEGA             & 15.19 & 15.19 \\
\midrule
\multirow{3}{*}{Reasoning} 
    & BigBenchHard      & 70.10 & \textbf{71.38} \\
    & ZebraLogic        & 0.00  & \textbf{20.90} \\
    & AGI Eval (EN)     & \textbf{73.91} & 73.75 \\
\midrule
IF & IFEval            & \textbf{82.25} & 72.07 \\
\midrule
\multirow{3}{*}{Knowledge} 
    & MMLU              & 77.91 & \textbf{80.16} \\
    & PopQA             & 22.64 & \textbf{26.31} \\
    & GPQA              & 32.58 & \textbf{38.61} \\
\midrule
Chat & AlpacaEval 2 LC  & \textbf{50.57} & 17.45 \\
\bottomrule
\end{tabular}
\caption{Performance comparison of 12B-parameter models. DictaLM-3.0-12B-Instruct shows strong knowledge retention (MMLU, GPQA) while Gemma 3 12B excels at mathematical reasoning.}
\label{tab:res-12b-models}
\end{table}

\begin{table}[h]
\centering
\small
\begin{tabular}{@{}ll rr r rr r@{}}
\toprule
& & \multicolumn{3}{c}{\textbf{Accuracy (\%)}} & \multicolumn{3}{c}{\textbf{Tokens}} \\
\cmidrule(lr){3-5} \cmidrule(l){6-8}
\textbf{Category} & \textbf{Benchmark} & \textbf{Inst.} & \textbf{Think} & $\mathbf{\Delta}$ & \textbf{Inst.} & \textbf{Think} & $\mathbf{\Delta}$ \\
\midrule
\multirow{2}{*}{Math} 
    & MATH       & 60.83 & 86.41 & \textcolor{bestcolor}{+25.6} & 2,233 & 1,457 & \cellcolor{green!15}$-$35\% \\
    & OMEGA      & 14.16 & 28.38 & \textcolor{bestcolor}{+14.2} & 2,319 & 5,145 & +122\% \\
\midrule
\multirow{3}{*}{Reasoning} 
    & BigBenchHard  & 64.12 & 73.00 & \textcolor{bestcolor}{+8.9}  & 535   & 669   & +25\% \\
    & ZebraLogic    & 23.60 & 47.80 & \textcolor{bestcolor}{+24.2} & 6,090 & 5,495 & \cellcolor{green!15}$-$10\% \\
    & AGI Eval (EN) & 73.12 & 82.93 & \textcolor{bestcolor}{+9.8}  & 650   & 994   & +53\% \\
\midrule
Inst. Following & IFEval & 87.80 & 88.17 & \textcolor{bestcolor}{+0.4}  & 666   & 2,230 & +235\% \\
\midrule
\multirow{3}{*}{Knowledge} 
    & MMLU       & 81.11 & 85.93 & \textcolor{bestcolor}{+4.8}  & 294   & 398   & +35\% \\
    & PopQA      & 27.83 & 30.59 & \textcolor{bestcolor}{+2.8}  & 33    & 794   & +2306\% \\
    & GPQA       & 41.96 & 55.13 & \textcolor{bestcolor}{+13.2} & 1,481 & 2,331 & +57\% \\
\midrule
Chat & AlpacaEval 2 LC & 31.11 & 74.11 & \textcolor{bestcolor}{+43.0} & 475   & 1,630 & +243\% \\
\midrule
\multicolumn{2}{l}{\textbf{Average}} & -- & -- & \textcolor{bestcolor}{+15.7} & 1,478 & 2,114 & +303\% \\
\bottomrule
\end{tabular}
\caption{Efficiency analysis: DictaLM-3.0-24B Instruct vs Thinking variants. The Thinking model achieves substantial accuracy gains across most benchmarks. Negative token overhead (green) indicates the Thinking model used \emph{fewer} tokens while achieving better accuracy.}
\label{tab:thinking-efficiency}
\end{table}

\section{Conclusion}

In this report we introduced \texttt{Dicta-LM 3.0}, a new suite of Hebrew-focused sovereign LLMs spanning 24B, 12B, and 1.7B parameters, each trained on large-scale Hebrew corpora and extended with long-context capabilities. Through continuous pre-training, supervised fine-tuning, and reinforcement-based post-training, the models achieve substantial gains across a wide range of Hebrew tasks, and maintain strong performance on English benchmarks despite the heavy Hebrew emphasis.

To address gaps in evaluating non-English chat models, we proposed a new Hebrew chat benchmark suite covering summarization, translation, reasoning, Israeli knowledge, and diacritization. The resulting evaluations show clear improvements over the base models and competitive performance against significantly larger multilingual LLMs. The 24B model, in particular, sets a new standard for Hebrew-capable open-weight models and demonstrates that sovereign LLMs can outperform generalist models when trained with focused data and targeted post-training.

Our work provides concrete methodology - data construction, training strategy, and evaluation tools - for adapting frontier LLMs to low-resource languages. We hope this effort supports broader development of sovereign language models and encourages further progress in multilingual NLP.

We are happy to release the models to the public with a permissive license:

\begin{itemize}
    \item \href{https://huggingface.co/dicta-il/DictaLM-3.0-24B-Base}{DictaLM-3.0-24B-Base} - 24B Base model 
    \item \href{https://huggingface.co/dicta-il/DictaLM-3.0-24B-Thinking}{DictaLM-3.0-24B-Thinking} - 24B chat model with reasoning and tool-calling capabilities. 
    \item \href{https://huggingface.co/dicta-il/DictaLM-3.0-Nemotron-12B-Base}{DictaLM-3.0-Nemotron-12B-Base} - 12B base model. 
    \item \href{https://huggingface.co/dicta-il/DictaLM-3.0-Nemotron-12B-Instruct}{DictaLM-3.0-Nemotron-12B-Instruct} - 12B chat model with tool-calling capabilities. 
    \item \href{https://huggingface.co/dicta-il/DictaLM-3.0-1.7B-Base}{DictaLM-3.0-1.7B-Base} - 1.7B Base model 
    \item \href{https://huggingface.co/dicta-il/DictaLM-3.0-1.7B-Instruct}{DictaLM-3.0-1.7B-Instruct} - 1.7B chat model with tool-calling capabilities. 
    \item \href{https://huggingface.co/dicta-il/DictaLM-3.0-1.7B-Thinking}{DictaLM-3.0-1.7B-Thinking} - 1.7B chat model with reasoning and tool-calling capabilities. 
\end{itemize}

\section*{Acknowledgments}

This project benefited greatly from contributions of numerous teams and institutions, without which this project wouldn't have been possible. 

We would like to thank the NVIDIA DGX Cloud Lepton team for early access to the platform, and for providing us with the necessary compute to train this model. We would like to extend our thanks to the technical team which made themselves available at all hours to make sure the training ran smoothly and assist in integrations with the platform: Noel Osagie, Greg Kosiorowski, Martin Piercy, and Anshul Jindal - thank you!

We would also like to give special thanks to the NVIDIA team, with whom we met regularly and who provided us with constant support and guidance throughout. Their extensive knowledge of the inner workings of every part of the training process - from training infrastructure to post-training methods - was an invaluable resource and a key part of our success. Specifically, we would like to thank Adam Henryk Grzywaczewski, Oleg Sudakov, Meriem Bendris, Ofir Zamir, and Asher Fredman for their significant contributions, along with Natalia Segal, Amit Bleiwess, Alyss Noland, Nader Khalil, Sergio Perez, and Liana Mikaelyan, whose expertise was instrumental in moving this project forward.

We would also like to thank Tal Geva and the DDR\&D IMOD / The Israeli National Program for NLP in Hebrew and Arabic for their invaluable contributions without which this project wouldn't have been possible.

\bibliography{anthology,custom}
\bibliographystyle{acl_natbib}

\newpage

\appendix
\section{Appendix: Lighteval command for base model evaluation on English tasks}
\label{appendix:leval-eng-command}

\begin{tcolorbox}[colback=gray!5!white, colframe=gray!80!black,
  sharp corners, boxrule=0.7pt, 
  top=2mm, bottom=2mm, left=2mm, right=2mm]

\begin{lstlisting}[basicstyle=\ttfamily\small, breaklines=true, columns=fullflexible, keepspaces=true]
lighteval vllm "model_name=google/gemma-3-27b-pt,tensor_parallel_size=2,max_model_length=32768,trust_remote_code=True" "helm|commonsenseqa|5,leaderboard|arc:challenge|25,leaderboard|winogrande|5"
\end{lstlisting}

\end{tcolorbox}

\end{document}